# Open Source User Mobility and Activity Datasets:
# Taxonomy and Applications

Sinjoni Mukhopadhyay King — simukhop@ucsc.edu
Katia Obraczka — katia@soe.ucsc.edu
Faisal Nawab — fnawab@uci.edu

***Abstract*—**The study of user mobility and activity (uMA) has a wide-range of applications, including network resource planning, connected healthcare, localization, social media, e-commerce, etc. The current state-of-the-art in uMA research has extensively relied on open-source traces captured from pedestrian, vehicular and application based activity. Most of these traces are feature rich and diverse, not only in the information they provide, but also in how they can be used and leveraged. This diversity poses two main challenges for researchers and practitioners who wish to make use of available uMA datasets, classify existing uMA datasets or create new uMA datasets. First, there is no readily available bird's eye classification of the existing open source uMA traces, which makes determining whether the identified datasets are adequate typically labor- and time consuming. Second, given a trace, it is very difficult to identify the key features and their use cases without performing a detailed analysis of the traces. The purpose of this manuscript is three-fold. First, we propose a taxonomy to classify open-source mobility traces by their mobility mode, data source, collection technology and application type. These taxonomy buckets can be used to create tags for existing and new datasets making the search for problem specific datasets much easier than current searches. Second we use the proposed taxonomy to show how existing datasets can be classified using the buckets, using examples of popular open source uMA traces, along with providing their publishing source, licensing, and anonymization strategy. We also briefly discuss how the taxonomy can be used to guide collection of newer uMA datasets. Third, we highlight three case studies using popular publicly available uMA datasets to showcase how our taxonomy can be used to tease out feature sets in traces to help determine their applicability to specific networking, health, lifestyle and location based use-cases.

**Index Terms—**Mobility Traces, Open Source, Taxonomy

## I. INTRODUCTION

Understanding patterns in user and vehicular movements and activity has important applications in a wide range of fields like urban planning [1], public health and healthcare [2], efficient transit and transportation [3], critical infrastructure planning and deployment [4], commerce, entertainment [5], network, data communication and connectivity [6], [7], among others.Several key stakeholders have acknowledged the exponential growth in mobile data traffic volume, i.e., data traffic generated by mobile net, as well as the wide range of applications of uMA datasets. For example, Ericsson's Mobile Data Traffic Outlook report [8] forecasts global mobile data traffic growth from a total of 65EB per month at the end of 2021 to 370E per month in 2027. The report in [9] highlights the need of coping with this unprecedented growth and still being able to provide adequate service to users. "Mobility as as Service" is discussed in [10] where public vehicular mobility analysis results are used to increase public transport convenience to the point where people consciously choose to use public over personal transportation. Arity, a company that focuses solely on mobility analysis, has their own blog outlining various applications of both vehicular and user mobility data [11]. These reports provide additional hard evidence and underscore the need for the availability and access to uMA datasets as vital to a wide range of critical applications, such as: (1) Adequate dimensioning of the underlying network infrastructure, including wireless and fixed broadband access network; (2) Scale and accommodate future connectivity and traffic demands as well as design systems and applications that are able to adapt to user mobility and activity patterns; (3) Urban planning such as transit, transportation, housing infrastructure, and emergency response (including public health emergency situations like the COVID-19 pandemic), as well as other services such as local government and community services, shopping, entertainment.

The importance of understanding uMA data has led to extensive efforts from academia and industry to collect [12], [13], analyze [14]–[17], and synthetically generate [18], [19] such data. These prior efforts, however, face some significant challenges as follows:

• ***Diversity in data***: uMA data is diverse to the point where small changes in the environment—e.g., the driving application, collection technology, geographic location, user demographics—can trigger significant change in the data. For example, existing taxi traces from Rome focus on temporal patterns using timestamps to categorize locations visited based on dates, while taxi traces collected in San Francisco focus on spatial patterns categorizing locations based on unique cab identifiers. This challenge places considerable burden in efforts that aim to improve access and analysis of uMA data.

• ***Data privacy and protection:*** uMA data is typically personal identifiable information (PII). This means that collecting, storing, processing, and sharing such data is restricted due to data privacy and protection regulations such as GDPR, CCPA, and HIPAA. User consent is needed to collect, store, and process this data, and sharing data needs to be performed in a manner that does not reveal any PII data. This significantly constrains access and collection of uMA data and has led organizations to not be able to share it without anonymization and summarization techniques, which in turn limit the information that can be shared and insights that can be derived.

(uMA) datasets include traces that track user mobility, network activity patterns, preferences, habits across different scenarios. Some examples include seasonal travel patterns, public transportation and network usage, communication between devices, health activity data etc. This manuscript tries to fill these gaps by: (1) Proposing a taxonomy that classifies publicly available uMA traces based on a number of factors including their uMA mode, data source, technology used for data collection, information type and their current and potential future applications; (2) Categorizing approximately 30 well known public datasets using our taxonomy; and (3) Analyzing three popular traces, each uniquely categorized according to our taxonomy.

Additionally, uMA datasets and models have been extensively used in the study of wireless networks and their protocols [20], [21]. Indeed, most network simulators include synthetic uMA generators, which, given a pre-specified uMA regime, determine the position of network nodes over time during simulation runs. More recent uMA models try to represent user mobility and activity more realistically by using real uMA traces, especially since an increasing number and variety of these real, open-source traces can be accessed through public services like CRAWDAD [12], Data World [13], Kaggle [22] and GitHub [23]. Most of these datasets are rich in information and have diverse applications. This poses two main challenges to researchers and practitioners who wish to make use of these open-source uMA datasets. First, it is quite difficult to get a bird's eye view of existing open-source traces without spending considerable time and effort searching and inspecting them. Second, once the traces are identified, determining whether they are adequate for the application at hand is far from trivial. These challenges motivate the need to have a framework that can systematically and meaningfully categorize uMA traces.

Existing surveys on uMA modeling focus mainly on two areas, namely uMA feature analysis techniques, where models with different feature characteristics are compared [24]–[29], and uMA prediction models [30]–[33]. They highlight different application areas like mobile and wireless networking, traffic modeling, smart city planning, etc. Additionally, the few more recent surveys that discuss uMA datasets focus only on traces belonging to a single category. For example, the survey presented in [34] compares uMA traces collected via Bluetooth sensors on two different university campuses. To the best of our knowledge, our work is the first to survey the current state-of-the-art in publicly available uMA traces, offer a taxonomy to put them in perspective, and provide a comparative analysis of existing traces according to the proposed taxonomy. Additionally, our manuscript is also meant to be a guide for researchers and practitioners to better navigate and leverage the space of currently available uMA traces as well as future ones.

**A. Taxonomy Overview**

Coming up with a representative taxonomy for uMA traces is not trivial due to their feature and application diversity. To create our main taxonomy we used a bottom up approach, starting from the data source or technology used to collect the traces, all the way up to the uMA mode i.e., pedestrian or vehicular, being represented by these traces. Another challenge we faced was finding a representative collection of traces to define our taxonomy. Most current state of the art uMA modeling techniques generate their own traces, only a fraction of which are put on public domain. Our strategy was then to select datasets that have been widely used with the goal of creating a classification scheme that is broad enough so that existing or new traces can be categorized using our taxonomy. Additionally, our study also analyzes potential applications of these traces to identify significant gaps in availability of real uMA traces, and thus motivates the need for realistic uMA traces generators.

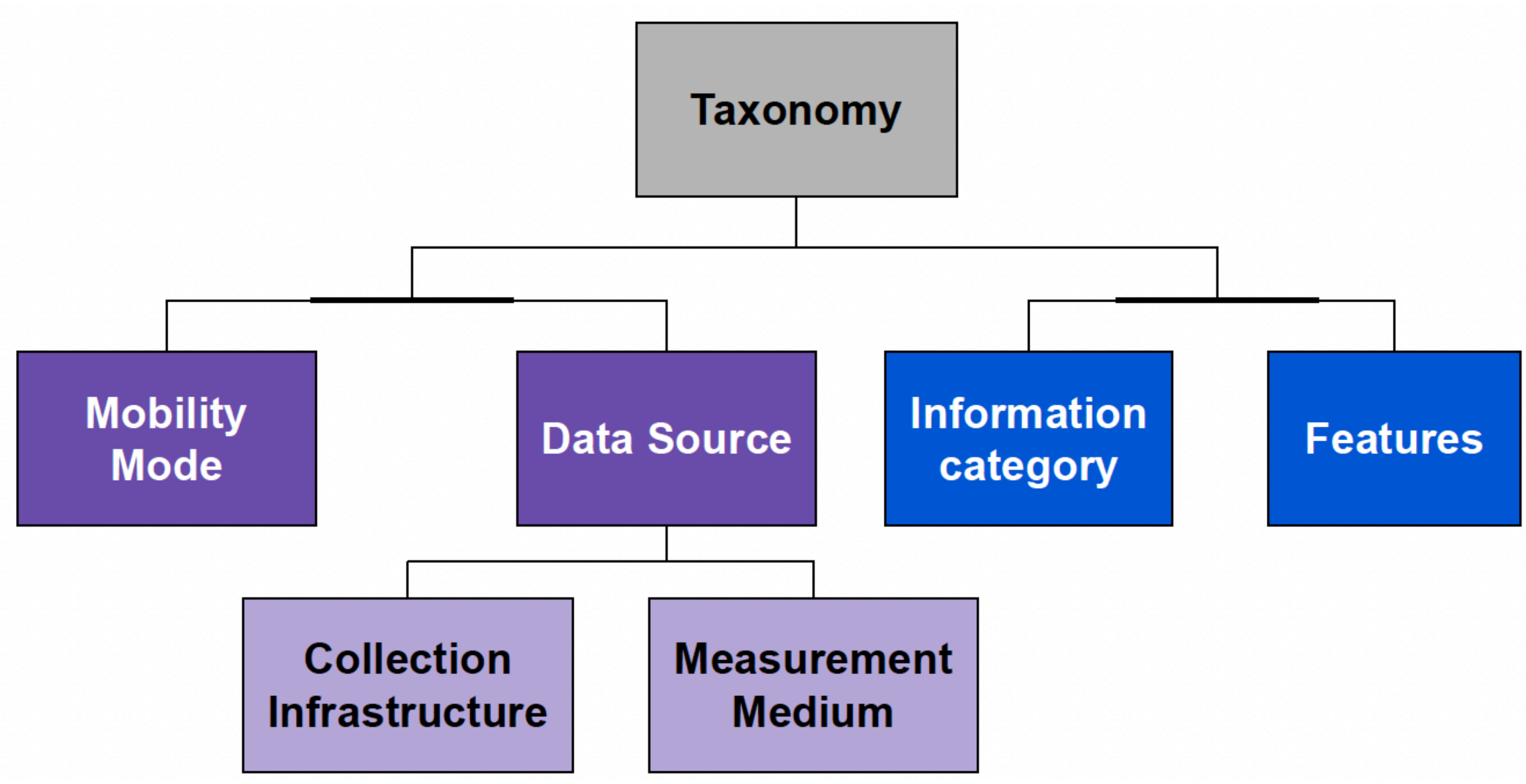

Figure 1: Taxonomy Overview

An overview of our taxonomy is illustrated in Figure 1. Under the left branch we have:

• **Mobility Mode**, which refers to the user's mode of transportation, namely pedestrian or vehicular.

• **Data Source** considers how the trace was collected and is further subdivided into:

– **Collection infrastructure**: systems that host the devices used to collect data.

– **Measurement Medium**: actual device/technology that generates the different measurements used to populate the datasets.

Pedestrian mobility typically represents movement within a limited geographic region, while vehicular mobility involves movement using various modes of transportation usually spanning larger geographic regions. Vehicular mobility includes personal vehicles and public modes of transportation like buses, trains, shared scooters/bikes/cabs, ships, airplanes, etc. In the case of pedestrian mobility, data is usually collected through smartphones, laptops, tablets, wearable devices or through network infrastructure gear. In vehicular mobility, data is either gathered through end-user devices like smartphones, laptops, tablets and wearable devices, in which case it is usually generated when these devices are inside a moving vehicle; or through smart-vehicle hosted devices. A more detailed example collection infrastructure and measurement media are illustrated in Figure 2. As illustrated in Figure 1, under the right branch we have:

• **Information Category**: Application groups created by studying existing open source mobility traces.

• **Features**: Raw and derived information types generated in open source mobility traces.

Based on the existing set of open-source traces, we have identified four main information categories:(1) Connectivity traces are typically used to optimize network performance, e.g., provisioning, redistributing resources to better manage network traffic, etc [35]–[46]; (2) Location traces can be used for location-based optimizations like improving waiting area around a specific business that has a heavy footfall [35], [37], [38], [41], [43], [44], [46]–[52]; (3) Health-related traces are applied to improve health solutions like adding features for call-for-help services; there aren't any current open source health traces because of HIPAA compliance, but we include this class as a future categorization possibility; (4) Lifestyle traces are used to draw patterns in user behavior like sleep cycles, downtime etc. Note that the taxonomy also lists, under every information category, some examples of features that may be present in the respective category of traces [48]–[50], [53]–[56].

### B. Roadmap

The rest of this document is organized as follows. Section I outlines challenges associated with open source data and Section II discusses the current state-of-the-art in uMA research. In Section III, we present our taxonomy to classify uMA datasets. Section IV categorizes 31 well-known public datasets using our taxonomy and discussing their applications. In the same section we analyze 3 uMA traces that represent distinct categories according to our taxonomy. Finally, Section V concludes the manuscript and discusses future trends in uMA research.

## II. USER MOBILITY AND ACTIVITY: LAY OF THE LAND

Better understanding of human activity and mobility in today's information-driven world has become increasingly essential in various areas such as network and communication infrastructure provisioning and deployment, urban planning, health care delivery, to name a few. In this section, we review existing efforts to study uMA by grouping them as follows: uMA feature analysis, uMA prediction, uMA surveys, and evolution of uMA dataset public accessibility.

### A. Feature Analysis of uMA datasets

Positioning and localization technologies like GPS, cellular tower based geo-positioning, WiFi positioning and other technologies used to track motion have enabled additional, more sophisticated approaches to collecting human mobility data and mining patterns of interest. Feature vector studies using uMA datasets were introduced in the late 1990's with the goal of analyzing human interactions in various environments that could provide information about cultural group formation [24].

Between 2010 and 2015, uMA feature studies took off again using GPS traces to mine geo-locations [25] and geocommunities [26]. There are also studies that focus on coarse- versus fine granularity of uMA datasets [27], location-dependent versus location-independent datasets [28], periodic transitions between locations affecting human activity [57], and citywide GPS logs from taxis [58]. Other references conduct comparative studies of different GPS-based trace analysis techniques [29]. Other kinds of uMA trace analysis involve data from location-based social network (LBSN) platforms to: extract and infer the purpose of travel, or the activity at the destination of a trip in daily life scenarios [59]; or study the impact of location history collection on uMA features [60]; investigate human movement among points-of-interests (PoIs) [61]–[63]; exploit information on transitions between types of locations, mobility flows between locations, and spatio-temporal characteristics of user check-in patterns [64]. Datasets captured from applications like Twitter are information-rich, e.g., they can indicate diversity in movement modes among individuals as well as movement within and between cities [65]. Some references also talk about using multiple sources of data from both cell phones and transit [66], and extracting uMA patterns using tensor decomposition techniques [67]. Others disc*uss inferring human activity patterns from anonymized mobile communication usage* [68].

In the last five years, with advances in data mining and data analysis techniques, several references have talked about the importance of Point of Interests (PoIs) and temporal distance to understand mobility patterns [69], using mobile and sensing data to analyze human habits and living environments [70], [71], mining human behavior and patterns from geo-socially tagged data [72], and learning mobility patterns with minimal user intervention [73]. With the rampant usage of deep learning (DL) techniques, DL-based feature extraction approaches have been used to analyze trajectory and transportation based mobility traces [74], [75]. Additionally, recent fog and edge computing technology has paved paths for healthcare [76] and transit [77] based mobility feature extraction.

More recently, user activity and mobility feature analysis has expanded to include Mobility-as-a-Service which allows paid access to mobility services like digital transportation management, environmental and health impacts of uMA patterns and choices, economic trends affected by social analysis etc. We divide these analysis techniques by application categories: connectivity, location, health and lifestyle.

**Lifestyle** based feature analysis included spatial, temporal analysis of geo-tagged Twitter data of Singapore residents to reduce crowds [78]; social media based opinion and pattern analysis to discover user mobility patterns, estimate polarized political opinions and tag interesting social media discussion topics [79]; improving energy efficiency in location by studying direct impact of building user mobility on operational and transport energy [80]; using social analysis to improve social exchange in uMA industry with respect to daily habits of adapting to connected vehicles, electrical motorization [81]; analyzing determinants of active mobility choices to compare the demographic, socio-economic and cultural factors that influence it [82].

**Health** based feature analysis included Mobility and Trajectory based Technique for Monitoring Asymptomatic Patients (MTT-MAP) [83] that used time-ordered spatial and temporal trajectory and uMA records of Asymptomatic patients towards reducing the stress of socio-economic complications in the case of pandemics; using Meta's user mobility database to identify the role of infection threats and containment policies, through labor commuting flows and business travels [84], spatial analysis of changes in urban uMA patterns and the modal distribution of transport to correlate with the evolution of environmental air quality indicators in the city of Spain [85].

**Location** based feature analysis included analysis of various service attributes of transportation modes (car-sharing, private car, and taxi) along with socio-demographic attributes of users, to optimize car-sharing strategies [86]; Mobility-on-Demand and Mobility-as-a-Service applied to fixed-route and on-demand service policies for low income communities [87]; Analyzing user mobility and activity patterns from GPS data to study differences between non-work/non-home locations of working/non working users on workdays/off days [88]; Using difference between prediction of trajectories among two different locations to optimize travel paths among the locations [89].

**Connectivity** based feature analysis included studying activity of a sensor device and its effects on high latency, which can result in low quality of services [90]; Privacy preserving, uMA supported federated learning vehicle algorithms [91].

### B. Modeling and Prediction using uMA data

uMA modeling in the 90's focused heavily on communication systems applications. Examples of uMA models include: using residence time distributions to analyze channel holding time [30]; using features of asynchronous point-to-point communication like distribution of processes to locations, routing of messages, failure to reach locations and their detection, to extract uMA patterns of processes [31]; and supporting activity in IPv6 without loss of connectivity [32]. uMA models studying mobility and activity of elderly people and their quality of life was also briefly studied [33]. uMA prediction between 2010 and 2015 started branching out into several new areas. Data from location based sensing networks, like a person's GPS trajectory, were used to predict current and future locations visited by users, how frequently they were visited [92]–[95] and to find additional points of interests [96]. Communication-based uMA modeling considered opportunistic networks and used data shared by short range devices to predict user communication patterns [97]; it also targeted uMA-aware personalization and resource allocation for mobile cloud applications [98]. Transportation-based uMA models use bus/taxi travel requests to predict bus travel demand for different routes as well as locations for potential future customers [99]. There has also been some psychology based human mobility and activity studies on regularity and predictability of human movements [28], [100], predicting uMA in response to a large-scale disaster [101], [102], and predicting long-term activity associating location information with contextual features like days of the week [103]. More recently, with uMA prediction riding the machine learning wave, there have been several references to DeepMove [104] that uses recurrent neural networks (RNNs) to predict human trajectory data, hidden Markov models to predict user movement [105], federated learning as a privacy-preserving mobility prediction framework [106], Deeptransport to predict user's future movements and transportation mode for a period of time [107], DeepUrbanMomentum for prediction of short term urban mobility [108], variational trajectory convolutional networks to predict point of interests [109], and Neural Turing machine with Stacked RNNs to predict neighborhood human mobility patterns [110].

More recent trends in mobility modeling and prediction have focused on topics like tourist choices, network resource scheduling, situation based activity trend prediction, edge computing optimizations etc., we try to highlight some of those topics again dividing them based on the information based taxonomy layers. **Lifestyle** based modeling and prediction included PredicTour which performs uMA modeling via social media profile extraction, to predict tourist activity [111]. **Health** based modeling and prediction included analyzing factors like number of new cases, social distancing, stay at home orders, domestic travel restrictions, mask-wearing policy, socioeconomic status, unemployment rate, transit mode share, percent of population working from home, and percent of older (60+ years) and African and Hispanic American populations, to predict user motion and activity within USA in the early days of the pandemic [112]. **Location** based modeling and prediction included analyzing spatio-temporal correlations and multi-type urban transition flows to predict individual traveling behaviors [113]; predicting the supply/demand of transport systems for efficient traffic management, control, optimization, and planning [114]; privacy-aware human trajectory prediction using adversarial networks [115]; group-based multi-features move (GMFMove), that constructs a uMA prediction model based on factors like the sequence of location, the category of location, and the geographic relevance of human mobility and activity [116]; Prediction by Partial Matching (PPM) to forecast each vehicle's path and cluster the vehicles with similar future path, moving direction, and moving speed into one group [117]. **Connectivity** modeling and prediction included optimizing system performance with wireless resource scheduling methods for high activity cases by predicting traffic volume [118]; propagation delay prediction using energy-efficient mobility based localization scheme [119]; Lightweight uMA prediction and offloading framework (LiMPO) that optimizes latency and energy consumption while improving the resource utilization of mobile edge computing servers [120]; MoSaBa, a wireless crowd charging method which leverages uMA prediction and social information for improved energy balancing [121].

### C. uMA Surveys

Over the years there have been various surveys outlining the state-of-the-art of user mobility and activity research. We have grouped the surveys by their application type. **Lifestyle** surveys, where Thorton et al surveyed user characteristics and their effect on the uMA, especially reactions to environmental change [122]; Barbosa et al surveyed geolocation data to study individual versus collective uMA patterns [123]; Lin et al surveyed data mined from GPS trajectory data focusing on locations significant for prediction of future moves, detecting modes of transport, mining trajectory patterns and recognizing location-based activities [29]. **Location** surveys for example, Palmer et al surveyed various gathering and analysis techniques for spatially-rich demographic data using mobile phones [124]; Asgari et al surveyed datasets representing population flow in transportation networks along with their data types and various applications [125]; Toch et al analyzed large scale uMA datasets using machine learning techniques [126] focusing on the data's positioning characteristics, the scale of the analysis, the properties of the modeling approach, and the class of applications. **Connectivity** surveys where Karamshuk et al analyzed challenges associated with uMA in Opportunistic Networks research and also reviewed uMA analysis and models [127]; Becker et al studied uMA characterization with respect to cellular network data [128]; Hess et al described steps for creation and validation of mobile networking based uMA *models [129]; Yang et al surveyed wireless indoor localization using inertial sensors [130].* **Other** uMA based surveys include Solmaz et al discussing commonly used metrics and data collection techniques for various models and also proposed a taxonomy to classify uMA models based on their main characteristics [131]; Wang et al surveyed uMA prediction models derived using multi-source datasets [132]. So far there have been no surveys on health specific uMA modeling.

From this summary of related work, it is clear that user mobility and activity is studied across an extremely broad scope of topics. However, to the best of our knowledge, our work is the first to provide comprehensive documentation and taxonomy classification of the various open source uMA datasets. Using our paper, the researcher's have access to a high level classification of a wide range of popular uMA traces along with granular details about their publishing source and privacy preserving properties. They can also use our taxonomy to guide classification of a newly generated uMA trace, which
will help them scope out potential application areas for the new trace.

### D. Evolution of uMA dataset public accessibility

Historically, uMA datasets have been open sourced through academic and open-source services like CRAWDAD [12] by Dartmouth, Google's subsidiary Kaggle [22], and Data World [13]. These services have a large trove of existing datasets, millions of subscribers, along with capability to add new datasets to their repositories. For example, Kaggle allows users to upload datasets as large as 100 gigabytes. They also provide data preprocessing on the raw datasets to make them more readable to the average user. Despite all of these benefits, there are two major challenges with these services. First, albeit free, they require users to register with the service, which means the service has access to user private information, i.e., name, e-mail address, etc. Second, locating uMA data within dataset repositories can be challenging, even using keyword search mechanisms. This usually results in users having to manually go through each dataset summary to pick the relevant dataset.
Other noteworthy examples of other uMA data open-source services include platforms such as the RFDatafactory [133], which allows users to access and create custom WiFi datasets. RFDataFactory currently has 31 datasets and 15 contributors. They provide APIs for data preprocessing, visualization, feature extraction, etc. Another example of an open-source dataset generation tool is Synthea [134], developed by the MITRE Corporation which is a synthetic patient data generator that models the medical history of synthetic patients.

## III. TAXONOMY

In this section, we describe in detail the proposed uMA taxonomy as illustrated in Figure 2, including the criteria used to classify open-source uMA traces. Note that arrows represent the relation between categories at the different levels of our classification.

### A. Mobility Mode

The Mobility Mode layer organizes traces into two major categories: pedestrian and vehicular.
1) **Pedestrian Mobility**: Pedestrian mobility datasets can provide an understanding of how people move in certain areas according to various aspects like traveling distance, locales where people tend to congregate, and trends related to places visited, when they are visited, and for how long [54], [55], [135], [136]. Lately, there has been a lot of emphasis on understanding crowd management, e.g., identifying attractors and detractors, determining and optimizing wait times in various situations, localizing congestion and bottlenecks in crowded localities, all of which provide invaluable insights on how people move in different places and situations. Other

Examples of applications that can benefit from better understanding of pedestrian mobility include reduction of crowdbased carbon-dioxide emissions, population density control in crowded areas, and optimization of city infrastructure. The COVID-19 pandemic has made the importance of being able to model pedestrian mobility patterns even more critical so that it can be used to perform contact tracing and manage public spaces in order to better enact local policies and restrictions.

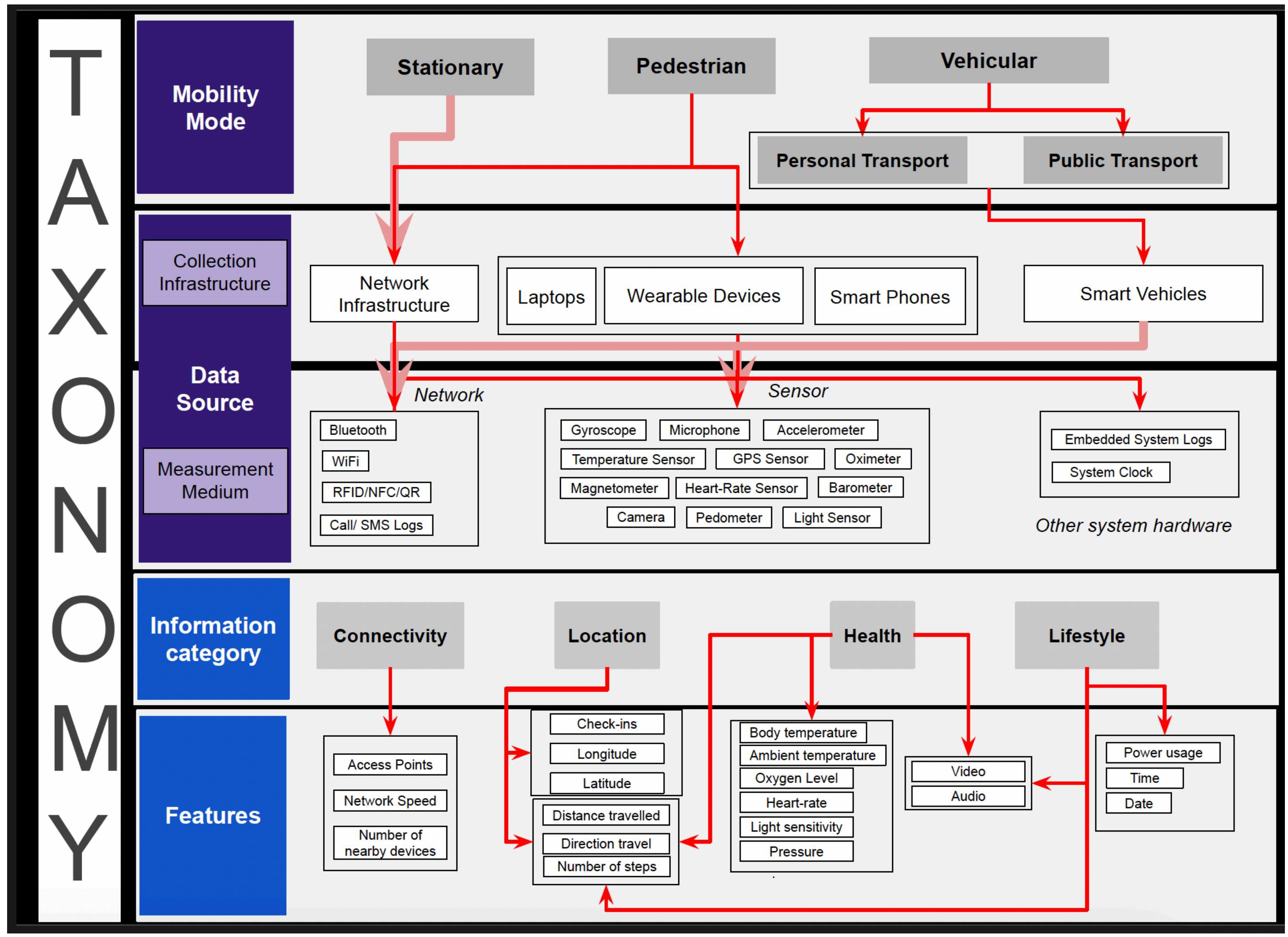

Figure 2: Mobility trace classification based on mobility mode and data source

2) **Vehicular Mobility:** Vehicular mobility traces can be used to characterize movement of various modes of transportation [47], [51], [52]. Before automobiles, cities were limited in terms of area, population and business prospects. The advent of different motorized transportation modes have fundamentally transformed the way cities are planned and expanded. Vehicular mobility can be broken down into personal and public transport. Public transport can further be differentiated based on mobility medium, namely: ground, sea, or air. Irrespective of whether publicly shared or personally owned, vehicular mobility captures various aspects of a given region like traffic during different times of the day, road congestion, favorable modes of transport, usage based on location specific infrastructure, etc. Understanding vehicular movement can also be insightful to understand trends among social communities. Other applications of vehicular mobility modeling include managing electrical vehicles (e.g., provisioning of charging stations), supporting autonomous driving, managing shared modes of transportation and supporting smart transportation services. Examples of vehicular mobility datasets include:

• **Personal Transport**, which is getting smarter by the day with new advances in 5G and IoT technology. With

these automobiles being connected to the Internet, there is a treasure chest of uploaded data that can help us model useful mobility patterns. Add to that application integration, where applications like Google and Apple Maps can record common routes and locations frequently visited, we can derive a complete picture of user mobility.

• **Public transport** by companies like Uber, Lyft, and Bay Wheels, which also collect mobility information via data shared by user applications. This includes information like pick up/drop off location, route taken, and stops along the way. Shared transport by local governments including buses and taxis which also provides us with information similar to privately-owned shared public transport.

**B. Data Source**

In the second layer, traces are then classified according to their data sources, including the infrastructure where data is being stored/located, and how data was obtained/measured.

**1) Collection Infrastructure**: Mobility data is typically collected by network infrastructure or end user devices. Network infrastructure includes devices that provide network connectivity to end users such as base stations and cell phone towers, that provide fingerprints which help identify times when a user has been around a certain location and for how long; as well as routers, switches, hubs, and wireless access points that provide network traffic exchange information which can be used to study mobility as a device switches between multiple components. End user devices can either directly be the source of mobility traces or traces can be extracted from applications hosted on these devices. Examples include:

- •Smartphones/Tablets/laptops that have become increasingly common as information and communication sources. They can collect/store information such as location, social media check-ins, communication logs like calls/messages being exchanged and sensor based information like motion tracking and health conditions.
- •Wearable devices, like smart-watches, with their sensing capabilities are ideal to monitor health conditions, and crowd density during ordinary and emergency scenarios.
- •Smart Vehicles that provide network and location information.

Pedestrian traces can be extracted from public network infrastructure and end user devices like smartphones, tablets, laptops, wearable devices, etc. Vehicular traces can be captured using data from specialized end user devices like transportation smart cards in personal and shared vehicles or, from transportation / transit applications or websites. Vehicular mobility can also be collected using sensors embedded in vehicles.

**2) Measurement Medium:** Here we consider how data was obtained or measured and divide traces into three categories: the Sensor category refers to traces collected via a variety of sensing devices; the Network category includes traces collected using devices that provide network connectivity (e.g., cellular, WiFi, Bluetooth, etc); and the Other system hardware group includes information derived from logs of on-device applications.

**Sensor-Based** data is usually the output from a device that measures the physical environment. The output of sensors is usually used as raw information or to trigger other sensors or processes. Below, we include examples of sensor data commonly found in mobility datasets.

• Global Positioning Systems or GPS sensors are receivers with antennas that use satellite based navigation to provide time and geolocation information usually in the form of latitude and longitude coordinates. In some cases, GPS sensors can also capture position in the form of velocity and orientation. These features are most commonly used as unique location identifiers. The datasets we have explored include latitude and longitude coordinates only.

• Light sensors are devices that convert any form of light energy, visible or infrared, into electrical signal outputs. In the case of mobility datasets, information on when it is day versus night can be useful to monitor patterns in user habits.

• An Accelerometer measures acceleration using three axes, X, Y and Z. Such sensors mainly provide two kinds of information: first, the static force applied on the sensor due to gravity and orientation; second, the force and acceleration exerted on the sensor in motion.
• Gyroscope sensors calculate angular velocity, or change in rotational angle per unit of time, usually measured in degrees per second. In the datasets we consider, gyroscope sensors add an additional dimension to the accelerator data to determine the orientation of a device.
• Magnetometers measure the relative change in magnetic field at a given location.
• Pedometers are mechanical devices that use software to detect vertical movement at the hip, to count the number of steps taken by a user. This can indirectly be used to derive information like distance traveled and patterns of other physical activities.
• Oximeters or pulse oximeters use LEDs to emit two types of red light through human tissue in order to measure oxygen saturation levels in the blood along with the number of times our heart beats per unit of time.
• Temperature sensors are electronic devices that measure surrounding ambient temperature and convert that into electronic data, to measure changes in temperature.
• Camera produces records in-terms of images or videos that can be used to derive social patterns in different human communities. This information can be derived based on the contents of the picture/video, location where the record was made, people who were a part of the record, what the people in the record were doing, etc.
Datasets containing sensor information can be collected directly by sensors and can be classified either under pedestrian or vehicular mobility.
**Network-Based** The ever increasing popularity and availability of mobile communications has made "anywhere, anytime connectivity" a reality. As such, end user mobility information that is collected through access network devices helps manage and provision network resources. Examples of network connectivity information contained in mobility traces include:
• Bluetooth technology targets short-range wireless communication.
• WiFi is one of the most widely used wireless technologies for data communication in local-area networks. It is also widely used as Internet access technology. Information from WiFi networks like access point associations / dissociations and signal strength can be used to determine user location as well as mobility patterns and trajectories.
**Other System Hardware** Mobility datasets can also include information generated by Application based data, also known as data collected from other system hardware, is actively triggered by users utilizing an application. Most of this data collected is only restricted to when the application is running and does not include information from the application's rest time. This kind of data usually gives information in the form of timestamps. Some specific datasets combine these timestamps with other information like location, human movement and constraints in the digital space. The Clock and the Calendar applications provide us with date and time, which can be useful if we want to model time series data, or analyze patterns and trends over specific time/date periods. The Map application uses requests to derive information about frequency of trips, locations in-terms of latitude and longitude, start and end times, types of transportation used in different locations for different times, pedestrian population in different areas of a city. Information generated by this kind of application can be useful for city planning and vehicle traffic management. Location based social network (LBSN) follows geosocial networking principles, where a social networking application has geographic capabilities like geotagging and geocoding to collect additional information about human social patterns. Location coordinates like latitude and longitude, added to uploaded pictures or social network check ins to cities, bridges the gap between the physical world and the online services, bringing social networks back to reality. Like the map datasets, this category of traces can also belong to the sensor group of datasets.

### C. Information Category

Open source mobility traces can also be grouped in terms of their application and features these traces contain. Information category can be roughly divided into four categories:

• **Connectivity**: Connectivity and network traces provide information on signal strengths, packet transfer details, network usage details and timestamps. Mobility performance metrics like user pause probability, user arrival, departure probabilities heavily impact the performance of 5G cellular networks. Optimizations can be performed by analyzing these metrics [137], [138]. Understanding user mobility characteristics, predicting network usage, can also help determine performance of routing protocols and feasibility of running an application over a vehicular ad hoc network [139]. Caching files based on popularity to reduce pressure on backhaul networks relies on user mobility pattern studies to provision storage allocation [140], model cost optimal device to device networks [141], [142] and improve data offloading [20], [143], [144]. Other applications of connectivity traces include analysis of spatial and temporal properties of pedestrian smart device based mobility datasets to enhance operations of wireless sensor networks [21].

• **Location:** Traditional location or GPS traces provide basic latitude-longitude information and when combined with location specific mobility tags, have several applications in various spheres of mobility analysis. Mobility tags can include any type of information tied to location like behavioral patterns, foot traffic, choices, network usage etc. Vehicular GPS information when combined with timestamps can provide insights on routes taken by specific modes of transportation at different times of the day. Frequency of travel using different modes of transportation at different locations can also be derived, which provide insights into mobility patterns of different communities. For example, if we have a trace that contains information of all buses along with their locations and timestamps for a particular city, we can identify the major hotspots or centrally located spots in the city based on locations that are visited most frequently by buses of different routes. GPS traces when combined with connectivity traces can provide insight into how well-connected some regions are. For example a trace that gives locations and their corresponding RSSI values can help identify placement of WiFi access points, which can in turn help offloading/managing network traffic within a specific location. GPS traces can also provide useful migration information. For example location information combined with unique person IDs, date and timestamps can help us identify patterns in how a person is moving between two locations.

• **Health:** Movement and health tracking traces include information from sensors like heart rate monitors, oximeters, accelerators and magnetometers. One important application of these traces is in the healthcare industry to derive conditions from health features. For example traces with information about a user's orientation and displacement can be used to predict whether the user is about to fall; this kind of information can be useful to enable independent living for older adults. Another important application of these traces is positioning and localization for users. For example navigation traces collected from sensors over time can help build a map for a particular area, complete with obstacles. This map can later be used for several applications like, video gaming using augmented reality, and accessibility applications like creating navigation tools for mobility-challenged users.

• **Lifestyle:** Geosocial traces, referred to as lifestyle traces in the paper, derive human patterns using social network check-ins. The unique style of geosocial mobility traces can provide insightful information for urban planning and retail real estate to property owners and operators. Geosocial data has the potential to reveal the personality of neighborhoods in a city. Building a park near neighborhoods that have a strong healthy living, nature or dog loving segments might be a source of support. Whereas building a shopping center in that same space instead might encounter resistance from the same group of people. Retail business success and geosocial segments are also closely related. A low priced, high traffic region may seem like a good place to build a store, but the most important factor contributing to the success of the store would be the social dynamics of the people around the store. Retail property owners can also use geosocial data to determine social segments of people around their property, which will help them lease the property to stores that are more likely to do well in the longer term in a particular area. Geosocial traces can also be used to target online applications to specific communities of people and be applied for marketing, merchandising and consumer goods. We can identify what people are doing and talking about in various locations. This kind of information can help with placement of billboards, local radio spots, or location-targeted mobile advertisements, as it is more effective to advertise in an area which has a social segment that has been predicted to be more receptive to those ideas. As a bonus these traces can be applied in planning healthcare facilities. Such data can be used to identify the age group of people in different locations, and based on the age determine if a particular location has more children who require pediatricians, or have an older population who require elder care physicians. We can also use such data to plan other specialized healthcare solutions, like chiropractors for locations where people have more of a sedentary lifestyle e.g. software engineers.

The features under these application brackets can be generated using the appropriate sensors, details of which we have already covered in the measurement medium section of the taxonomy. One thing to keep in mind when using the application based taxonomy trajectory is that due to the vast diversity in features within each application, there can be traces that belong to a subset of application categories instead of a single category.

## IV. USING THE TAXONOMY

As previously pointed out, our taxonomy can be used in a variety of ways. In this section, we illustrate some of the applications of our taxonomy, e.g., dataset selection (Section IV-A) and dataset collection (Section IV-B). Additionally, in Sections IV-C and IV-D, we use two datasets to demonstrate how our taxonomy can guide users to tease out three main sub-parts. The first subsection highlights the usage of our taxonomy as a guide to data selection for various problems and re-enforces the usage by using our taxonomy buckets to categorize existing popularly used uMA traces. The second subsection, as an extension of the first subsection, guides data collection of future datasets. In the end our taxonomy is a stepping stone to creating uMA specific data generation and storage frameworks. The third part of the section focuses on one pedestrian and one vehicular trace example, to showcase how our taxonomy can be used to tease out the relevant features of uMA datasets.

### A. Problem-based data selection

Analyzing uMA traces have a multitude of applications including communication infrastructure design and provisioning, urban planning, vehicular and network traffic analysis, community and public health services. Our taxonomy can be used to select datasets to be studied and analyzed to address specific problems or applications. For example, let's consider an urban transportation researcher who wants to study patterns in vehicular traffic across the different cities of the United States. The taxonomy provides the appropriate set of keywords to be used when searching for relevant datasets. They can shortlist a set of traces using high-level keywords like the vehicular mobility category, the data source GPS sensors or smartphones/smart vehicle sensors, the location application type. They can also use more detailed taxonomy keywords like features for the trace, timestamps and latitude, longitude, location information that can be used to derive trajectories. Table I illustrates how our taxonomy can be used to classify a set of popular uMA traces to highlight some of their main features. We discuss these traces and their features below.

T-drive [47], a vehicular dataset by Microsoft Research containing GPS trajectories, i.e., longitude and latitude of approximately ten thousand taxis in Beijing, has been used to derive optimized travel times, route and traffic prediction [154], [155]. This trace has also been used for non mobility related applications like testing database management systems and large-scale data processing techniques [156], and
testing density-based spatial clustering of applications with noise [157]. Other vehicular traces include Cabspotting [147] consisting of GPS latitude and longitude information collected from over 500 taxis; and mobility data collected from taxi cabs in Rome derived from GPS coordinates [51]. Such traces have been used to study mobility patterns in urban taxis [158], route regularity [159], social analysis in vehicular ad hoc networks [160], and identify outlier mobility trajectories [161]. Other less known vehicular traces that contain sensor information are the GRID bikeshare dataset, which describes the main attributes of GRID temperature, including the feed, operator, hours, calendar, regions, pricing, alerts, stations, and bike status [56]; and the mobility dataset from the city of Austin, Texas which includes GPS sensor information from bicycles and other means of transportation [149], [151], [162]. More recent taxicab datasets like the Chicago cabs trace collected location data from seven thousand licensed cabs operating within city limits has been used to route prediction and optimization [52].

Crivello refers to a pedestrian trace which includes sensor information from wearable devices and network connectivity information from smartphones; this trace has been used to compare and evaluate indoor localization solutions [35], [163], and for health applications like sleep quality monitoring [164]. Microsoft's Geolife dataset [136] consists of approximately eighteen thousand GPS trajectories with a total distance of 1,292,951 kilometers and a total duration of 50,176 hours collected from GPS loggers and GPS-enabled phones. The Geolife dataset has been used for transportation related applications like identification of transportation modes to create sophisticated intelligent transport systems [165]–[167], prediction of transport mode choices [168]; Privacy related applications like secure data compression in cloud [169], privacy preserving location preferences [170], [171]; and benchmarking performance of large datasets on prediction and generation models [172]–[174].

TABLE I
POPULAR uMA TRACES CLASSIFIED USING OUR TAXONOMY

| **TRACES** | | **EVALUATION** | | **APPLICATION** |
|---|---|---|---|---|
| | **Mobility Mode** | | **Data Source** | **Information category** |
| | | **Collection Source** | **Measurement Medium** | |
| T-Drive [47] | Vehicular | GPS | Sensor | Location |
| Crivello [35] | Pedestrian | Wearables/Smartphones | Sensor/Network | Connectivity/Location |
| Apple Maps [48], [49] | Pedestrian | Smartphones | Sensor/Other | Location/Lifestyle |
| Google Maps [49], [145] | Pedestrian | Smartphones | Sensor/Other | Location/Lifestyle |
| Descartes Lab [36] | Pedestrian | Smartwatches/phones | Sensor | Connectivity |
| KCMD-DDH [50] | Pedestrian/Vehicular | GPS/Smartphones | Sensor/Other | Location/Lifestyle |
| JRC(Europe) [53] | Pedestrian/Vehicular | GPS/Smartphones | Sensor/Other | Location/Lifestyle |
| GIM [146] | Pedestrian/Vehicular | GPS/Smartphones | Sensor/Other | Location/Lifestyle |
| Gowalla [55] | Pedestrian | Smartphones/Laptops | Sensor/Other | Location/Lifestyle |
| Brightkite [55] | Pedestrian | Smartphones/Laptops | Sensor/Other | Location/Lifestyle |
| Nsense [135] | Pedestrian | Smartphones/Laptops | Sensor/Other | Location/Lifestyle |
| Cabspotting [147] | Vehicular | GPS | Sensor/Other | Location |
| Geolife [136] | Pedestrian | Smartphones/Laptops | Sensor/Other | Location/Lifestyle |
| NYC Mobility [148] | Pedestrian/Vehicular | Smartphones/Laptops | Sensor/Other | Location/Lifestyle |
| GRID Bikeshare [56] | Vehicular | GPS | Sensor/Other | Location |
| Texas Mobility [149] | Pedestrian/Vehicular | GPS, Smartphones | Sensor | Location/Lifestyle |
| UILM [150] | Pedestrian | Smartphones | Sensor/Other | Location/Lifestyle |
| GSMC [37] | Pedestrian | Smartphones | Network/Other | Connectivity/Lifestyle |
| Flexran [38] | Pedestrian | Smartphones | Network/Other | Connectivity/Lifestyle |
| KTH [39] | Pedestrian | Smartphones | Network | Connectivity |
| BLEBeacon [40] | Pedestrian | Smartphones | Network | Connectivity |
| HYCCUPS [41] | Pedestrian | Smartphones | Network/Other | Connectivity/Lifestyle |
| Cambridge Haggle [42] | Pedestrian | Smartphones | Network/Other | Connectivity |
| Fire Dpt Asturius [43] | Pedestrian | Smartphones | Sensor/Network | Connectivity/Lifestyle |
| SocialBlueConn [44] | Pedestrian | Smartphones | Network/Other | Connectivity/Location |
| Rome Taxis [51] | Vehicular | GPS | Sensor/Other | Location |
| SIGCOMM 2009 [45] | Pedestrian | Smartphones | Network/Other | Connectivity |
| Commercial Seoul [46] | Pedestrian | Smartphones | Sensor/Network/Other | Connectivity/Lifestyle |
| Chicago taxi [52] | Vehicular | GPS | Sensor/Other | Location |
| Pedestrian Louisville [151] | Pedestrian | Smartphones | Sensor/Other | Location |
| MHealthDroid [152] | Pedestrian | Smartphones | Sensor/Other | Lifestyle |
| NetAAL [153] | Pedestrian | Smartphones | Sensor/Network | Connectivity |

Application generated traces like COVID traces derived from request information in Google and Apple maps [48], [175] consists primarily of location information represented by countries, regions, sub-regions and cities, combined with lifestyle information like transport type and location category. Other application based traces include data from Location Based Social Networks (LBSNs) like Gowalla [54], Nsense [135] and Brightkite [55] which use social network check-ins as the main source of the mobility data. We will elaborate on their applications in the next subsection. Application based traces can also include census and migratory information. Examples include the US Internal Lifetime Mobility (UILM) [150], which predicts mobility based on when and where a user is born and where they are currently located. The NYC Citywide Mobility survey of the New York City residents' travel choices, behaviors, and perceptions [148] collects mobility information via online surveys and phone surveys. The Knowledge Center on Migration and Demography, Dynamic Data Hub (KCMD-DDH) contains global transnational mobility data that provides us with information on country-to-country cross-border human mobility using global statistics on tourism and air passenger traffic [50]. The Knowledge Center on Migration and Demography, Dynamic Data Hub also highlights information on monthly air passenger flows, which can be synthesized into a set of indicators between countries worldwide; demography and mobility data collected by the Joint Research Centre (JRC) and the Directorate General for Regional and Urban Policy in European metropolitan regions in 2018 [53]; and region based mobility data collected via interactive maps publicly available on the Flows to Europe Geoportal, which provides statistical updates on migrant and refugee land and sea arrivals and routes towards Europe [146].

TABLE II
FEATURE LIST FOR POPULAR uMA TRACES CLASSIFIED USING OUR TAXONOMY

| TRACES | FEATURES |
|---|---|
| T-Drive [47] | Identifiers, Date, Time, GPS coordinates |
| Crivello [35] | IMU sensor data, WiFi and geo-magnetic field fingerprints |
| Apple Maps [48], [49] | Date, Time, Location, Transportation type, Usage |
| Google Maps [49], [145] | Date, Time, Location, Percentage increase/decrease in number of location visits |
| Descartes Lab [36] | travel information, dates |
| KCMD-DDH [50] | statistics from air passenger traffic and tourism |
| JRC(Europe) [53] | land and sea arrival informtion via travel portals |
| GIM [146] | land and sea arrival informtion via travel portals |
| Gowalla [55] | Latitude, Longitude, social network checkins, edge relations |
| Brightkite [55] | Latitude, Longitude, social network checkins, edge relations |
| Brightkite [55] | Latitude, Longitude, social network checkins, edge relations |
| Nsense [135] | Latitude, Longitude, social network checkins, edge relations |
| Cabspotting [147] | Taxi id, Date, Longitude, Latitude, Fare |
| Geolife [136] | GPS trajectories |
| NYC Mobility [148] | Survey for travel choices, user behavior |
| GRID Bikeshare [56] | GRID temperature, date, bike usage status |
| Texas Mobility [149] | GPS coordinates from different modes of transportation |
| UILM [150] | Census information, birth place, current home city |
| GSMC [37] | Device bluetooth encounters |
| Flexran [38] | Device bluetooth encounters |
| KTH [39] | User associations to their WiFi networks |
| BLEBeacon [40] | Device bluetooth encounters |
| HYCCUPS [41] | usage statistics, user activity, battery statistics |
| Cambridge Haggle [42] | Device bluetooth encounters |
| Fire Dpt Asturius [43] | WiFi, bluetooth, GPS information |
| SocialBlueConn [44] | Facebook friendships and interests |
| Rome Taxis [51] | Taxi id, Date, Longitude, Latitude |
| SIGCOMM 2009 [45] | Bluetooth encounters, opportunistic messaging, and social profile |
| Commercial Seoul [46] | GPS infor- mation, Wi-Fi fingerprints, user-annotated location information |
| Chicago taxi [52] | Taxi id, Date, Longitude, Latitude |
| Pedestrian Louisville [151] | GPS Trajectories |
| MHealthDroid [152] | IMU sensor, Physical Activity labels |
| NetAAL [153] | RSSI signatures, Room switch, Paths taken |

Another important consideration when studying uMA datasets is how they were collected, i.e., what kind of collection infrastructure and measurement medium were used. For example Bluetooth networks provide information about nearby devices and their characteristics [45], low energy packets generated by BLE beacons from end user devices like smartphones and laptops also provided user mobility information [40], the Cambridge Haggle dataset that contains bluetooth encounters between 12 nodes for approximately 6 days [42]; Asturias (Spain) Fire Department mobility and connectivity traces generated by GPS devices embedded mainly in cars and trucks, but also in a helicopter and a few personal radios [43]; traces containing Bluetooth encounters, Facebook friendships and interests of a set of users collected through the SocialBlueConn application at the University of Calabria [44].

Other notable network traces, also collected using applications, include the Global System for Mobile Communications(GSMC) which gathers information from approximately 10 mobile smartphone (iPhones) users via the MySignals iPhone App [37]; data collected by Flexran from a platform for software-defined radio access networks [38]; data collected using the HYCCUPS Tracer, that contains availability and mobile interaction information such as usage statistics, user activity, battery statistics, or sensor data, a device's encounters with other nodes or with wireless access points [41]; and traces with Bluetooth encounters, opportunistic messaging, and social profiles of 76 users, collected using the MobiClique application at the SIGCOMM 2009 [45]. Mobility data collected by organizations include records of authenticated user associations to their WiFi networks [39] and fine-grained network mobility data from commercial mobile phones in Seoul, Korea, containing continuous GPS information combined with Wi-Fi fingerprints and user-annotated location information [46]. In the following subsections we apply our taxonomy to three traces from this non-exhaustive list followed by some mobility analysis and outline. The three traces we are choosing are the COVID mobility traces by Google, since they are of the pedestrian type, the Cabspotting traces, since they are of the vehicular type, and the Brightkite traces, since despite being a feature-sparse dataset, they have several important applications.

**B. Guiding data collection**

The ever increasing popularity of mobile applications and services has fueled the need to access a diverse range of uMA datasets that can be used by researchers and practitioners in different disciplines ranging from communication infrastructure engineering to urban planning. As efforts to collect and generate uMA datasets proliferate, they can use our taxonomy to guide their data collection and generation activities. Using the previous example where a user wants to learn vehicular trajectory patterns across the US, they can use the taxonomy in two ways. First, they can look up existing datasets that contain keywords like ”vehicle/vehicular” or application keywords like ”location” etc. Then they can use the features of those datasets as a guide for what features to collect for their use case. In the case of the above user, they will come across a subset of data, containing datasets like T-Drive, Chicago, SF taxi traces, etc. Second, the taxonomy can also guide them on the data source used to collect the information. For example, the above user will know that the relevant data can be collected using smartphone sensors or GPS devices attached to the vehicle itself.

**C. Situational COVID analysis using data derived from mapping platforms**

Given the current ongoing battle the world is fighting against COVID-19, we apply our taxonomy to classify the very well known Google's COVID Mobility trend dataset, as shown in Figure 3, derived using data from applications like Google maps on smartphones. Companies like Microsoft, Google and Apple have extracted data from applications like Google and Apple Maps to analyze changes in mobility trends since the COVID-19 pandemic started in late 2019 [49]. We classify this under pedestrians since the information is requested on an application in a pedestrian smartphone. The raw maps dataset provides date-wise GPS locations and a percentage increase or decrease in mobility for various categories within each location. The categories include retail/recreation, grocery, parks, residential, workplace, and transit stations and are derived from location tags present in the map settings. The Google dataset, lays emphasis on public businesses and properties signaled in the requests. As such, we use it to analyze trends based on the increase or decrease in number of requests for the specific classes of locations/businesses. Figure 4 illustrates a monthly change in different map requests for the USA. This image was generated to visually represent data provided in the Google Mobility trend dataset. We can generate such charts for cities, states and/or at country levels. From the figure we observe that, as the months progressed from February to May, with increase in COVID threat, there has been up to a 38% decrease in visits to workplaces, a 35% decrease in retail and recreation Maps requests, a 30% decrease in public transportation usage, and a 25% increase in park visitations.

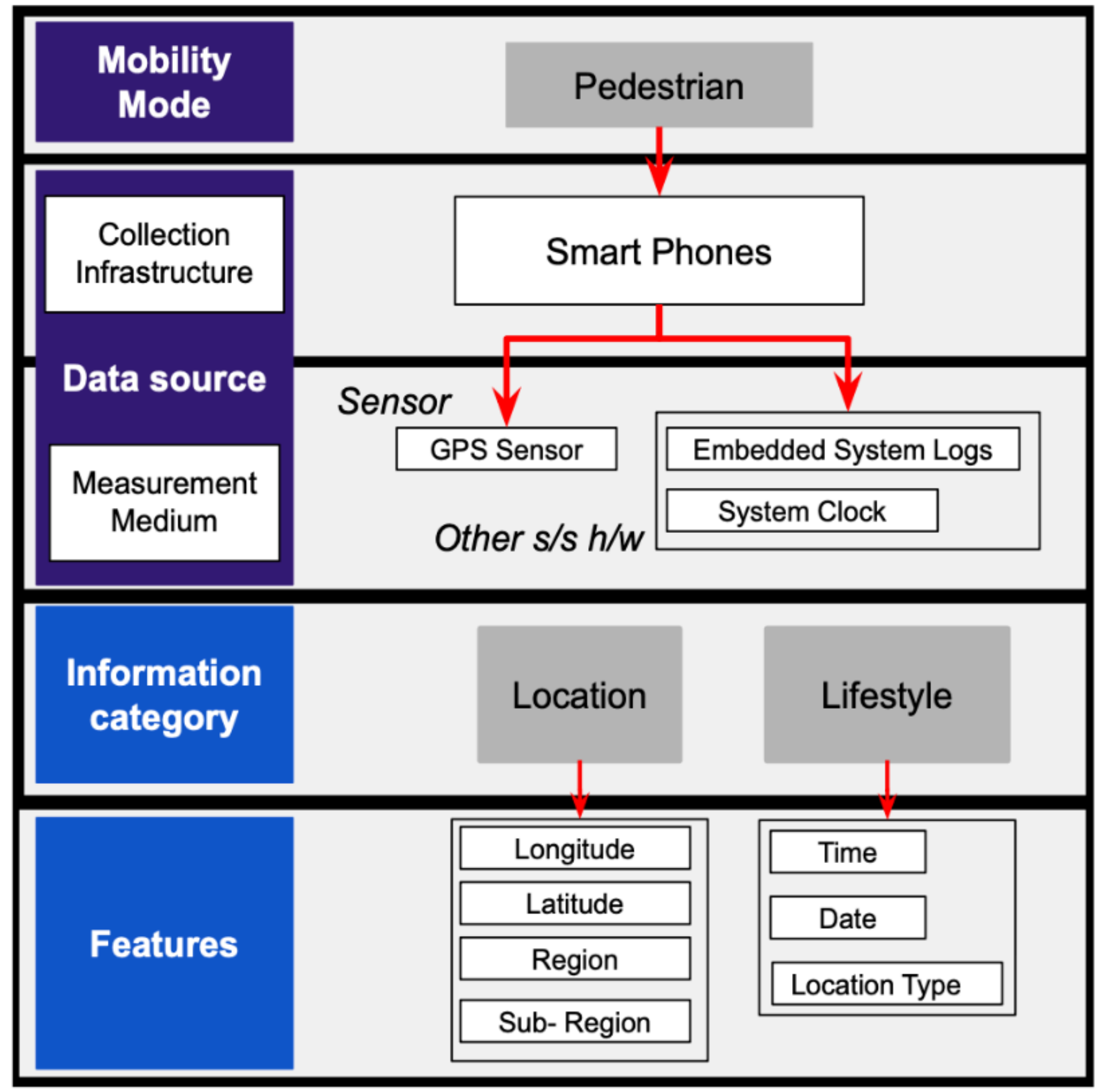

Figure 3. Classifying COVID Mobility Traces extracted from Google Maps

Real life applications of these traces are:

• **Lifestyle:** One real life application of this dataset from 2022 is combining it with the geo-spatial analysis of tweets in Singapore [78] to determine user visitation and amenities usage patterns at locations like parks, public links between parks and malls, taxi stands, residential areas, and shopping malls; Examination of the impact of early evening curfew on mobility by studying a shift in curfews from 9pm to 6pm in Greece using Google mobility data [176]; Studying the impact of mobility restriction strategies in the control of the COVID-19 pandemic to model the relation between COVID-19 health and community mobility data [177]; Using Google mobility reports combined with geotagged Twitter data to extract spatiotemporal human mobility patterns during this COVID-19 pandemic in New York City [178]; Description of the economic activity using the internet during COVID-19 pandemic, aimed to show the relationship between the people's mobility during COVID-19 with economic activity using the internet [179].

• **Health:** Other real life applications for the Google Mobility Report include: mobility used as a representation for risky behavior, comparing exposure before and after mask mandates were imposed [180]; Study the impact of mobility restriction on reducing the COVID-19 effective reproduction number in the Kingdom of Saudi Arabia [181]; Assess the impact of contact tracing in middle-income countries to provide data to support the expansion and optimization of contact tracing strategies to improve infection control [182]; using Google mobility report to support government policies, national culture and promote social distancing during the first wave of the COVID-19 pandemic [183]; Comprehensive survey of human mobility open data that can guide researchers and policymakers in conducting data-driven evaluations and decision-making for the COVID-19 pandemic and other infectious disease outbreaks [184];

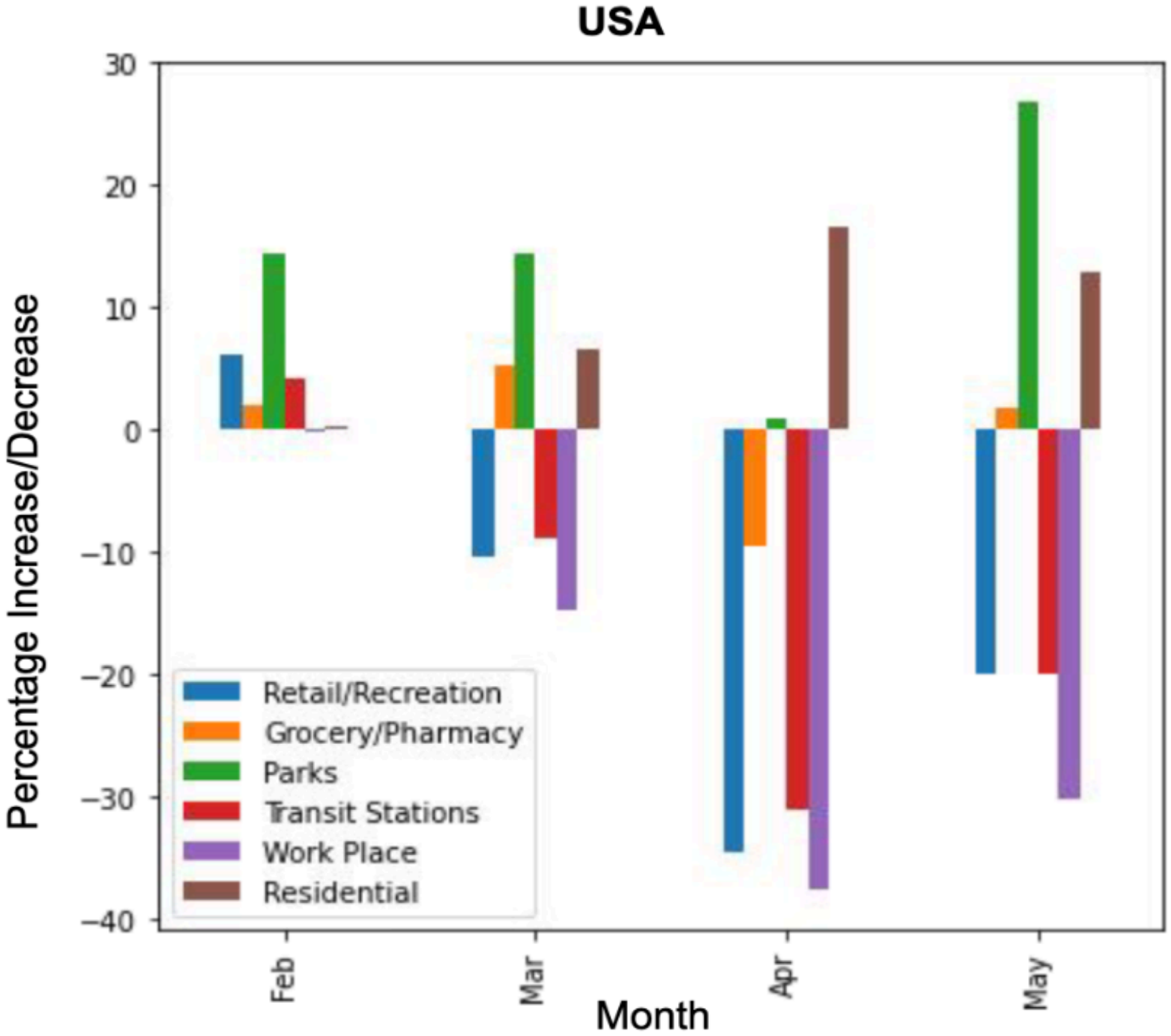

Figure 4. Mobility trends in businesses during COVID derived from information extracted from Google Maps for the year 2020.

• **Location:** Combining data from google report and local agencies to compare the transport impacts of the COVID- 19 in Germany and State of Qatar, based on the rates of infection and response measures [185]; Analysis of impacts of working from home on activity-travel behavior during the pandemic [186]; Combining Google mobility data and Apple maps data to track changes in community mobility and transport modes during the COVID-19 Alert levels [187]

• **Connectivity:**Evaluation of the association between social distancing quantified by mobile phone data and the current prevalence of COVID-19 infections in the U.S. per capita [188].

**D. Location based mobility pattern study using GPS data**

We highlight mobility patterns in two different datasets:(1) Derived from GPS traces from taxi cabs in San Francisco and (2) Derived from Brightkite, a location based social network application. Both datasets can be analyzed using the same mobility metrics. However, the taxi trace represents vehicular mobility patterns while the social network trace represents pedestrian mobility patterns. Figure 5 shows how the SF taxi dataset is classified using our taxonomy: It is a public transportation based vehicular trace. The data is collected using end user devices like smart GPS devices containing GPS locations, in the form of latitude-longitude tuples, and time/date information from the system clock. The San Francisco taxi trace associates each taxi with a unique taxi ID and a series of locations it visited over a period of time. The overall goal of the dataset is to provide locations that are commonly visited by the taxis in each region. The traces focus on spatial patterns categorizing locations based on unique cab identifiers. Such dataset graphs can provide us with various types of information. Figure 6 shows how the Brightkite traces are classified using our taxonomy. The dataset is pedestrian based, collected using applications running on end user devices like smartphones, iPads, or laptops. GPS location, region- and check-in information are collected and used to identify social relations between various groups of people. The technology type used provides information about visited locations, along with their corresponding latitude, longitude, and timestamps.

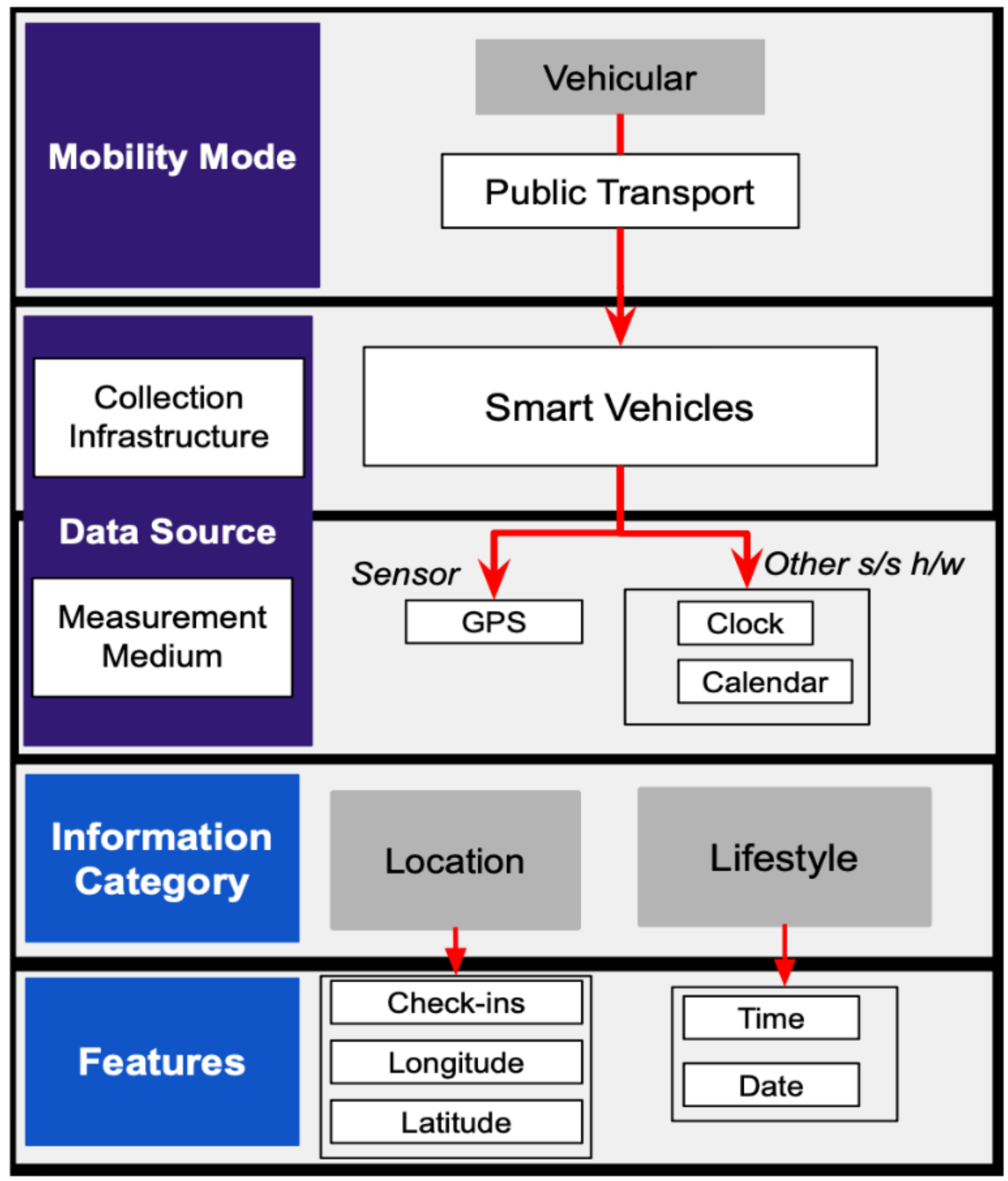

Figure 5. San Francisco GPS Taxi traces

We use some commonly used mobility metrics to highlight the different kinds of information that can be derived from these datasets. These metrics can be divided into two categories:
Collective metrics, which provide aggregated information about a group of users, and Individual metrics, which provide information on a per user basis:

• **Collective Metrics:**

– Number of homes in a given location, which can provide information about the neighborhood. A higher number of homes in a location could point to the area being residential.

– Mean Square Displacement is the measure of deviation of a set of users from an original point over time. This information can be used to determine the area a specific business covers e.g. a cab service running in a specific neighborhood of a city.

– Random location entropy: is the degree of predictability of a location, assuming each user visits it with equal probability.

– Uncorrelated Location Entropy: also known as temporal-location entropy, is the historical probability that a user visited a location.

– Visits per location describes the total number of visits made to a location by a set of users. The location based collective metrics can be used to determine the popularity of a location, when combined with date and time information, can help identify seasonal patterns in visits to locations. For example a beach town would have more visits during summer versus a hill station would have more visits in the winter months.

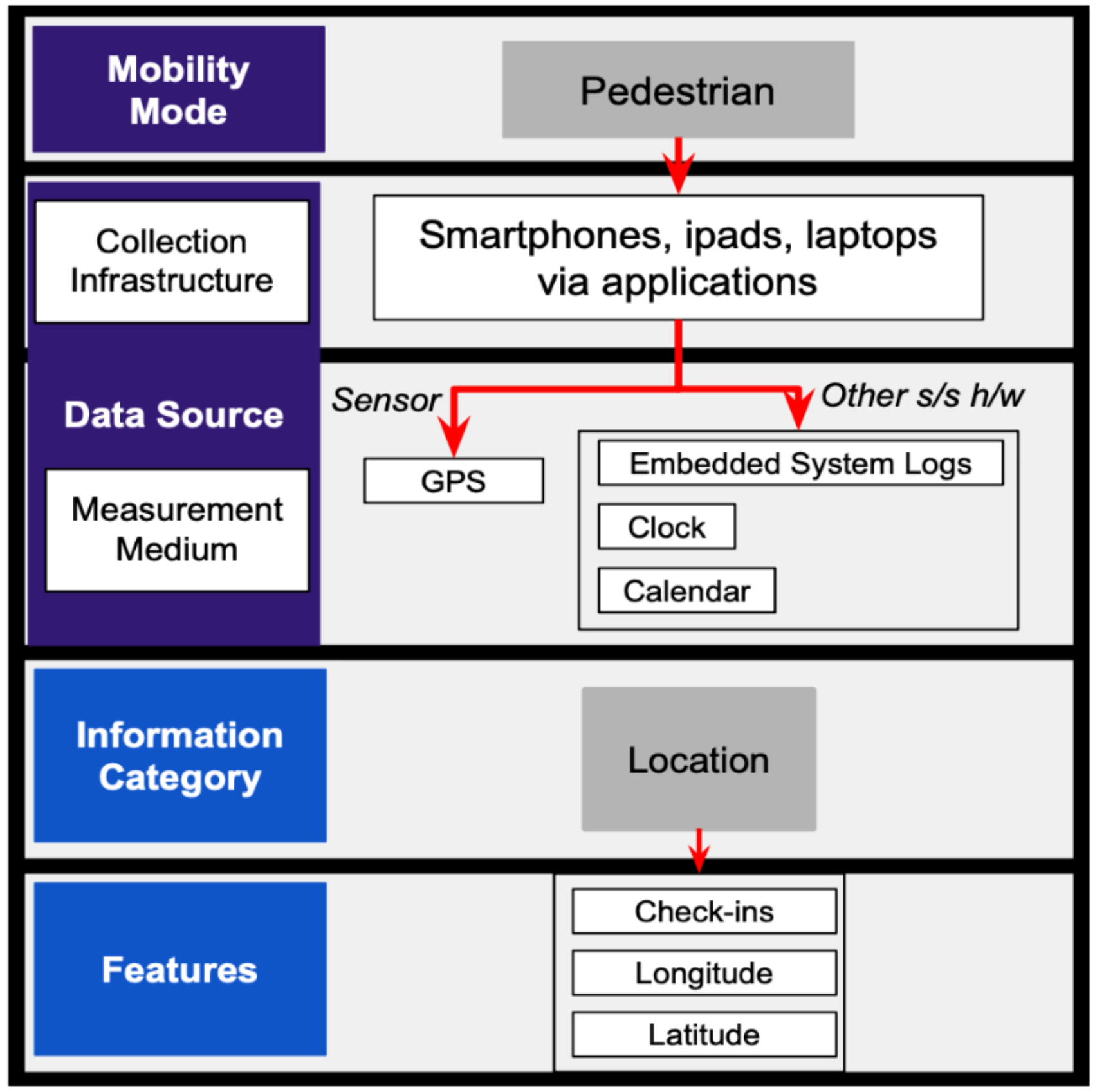

Figure 6. Brightkite Location Based Sensor Network Traces

• **Individual Metrics:**

– Straight line distance is the shortest distance traveled by a user between two points that does not contain any turns.

– Frequency rank of locations visited by an individual. This can be used to measure a user's choice of locations. For example, for a user whose top five frequency ranked locations are San Francisco, San Diego, Santa Barbara, Los Angeles, Santa Cruz, we can infer that they prefer visiting the West Coast over the other parts of the country.

– Home location gives the GPS coordinates of the exact location of a user's home. This information, when combined with the other metrics can help us determine patterns like how far a user travels regularly versus during the holiday season from home.

– Individual mobility network (IMN) is a set of locations visited by an user as represented by a graph with V nodes and E edges.

– Jump Lengths in kilometers is the geographic distance between two points traveled by an user

– Location frequency represents the visitation frequency of different locations for an user

– Maximum distance from home gives information on the farthest a specific user has traveled with respect to their home location.

– Maximum distance traveled by an user in kilometers

– Number of locations visited by a user gives the total number of locations that a user has visited during a given time period.

– Number of visits made an user gives the total number of visits across all locations that a user has made during a given time period.

– Individual radius of gyration represents the characteristic distance traveled by a user away from its center of mass.
– Random and Uncorrelated Entropy as defined previously but for the case of individuals instead of locations.
– Waiting times between movements of individuals. This metric is specific to taxi cabs and can help determine the average time that a cab had to wait for a rider.
Numbers like locations and visits can provide information about user travel patterns e.g. if a user likes to stay at home more often than travel, their number of locations visited will be lower than a user who travels to new locations often. Not every metric is applicable in all use cases.

TABLE III
RESULTING MOBILITY METRICS FOR SF TAXI TRACES FOR TWO DIFFERENT TAXICABS.

| Mobility Metric | Cab 1 | Cab 2 |
|---|---|---|
| Mean Square Displacement | 96.334 | 57.268 |
| Home Location | $-122.38, 37.61$ | $-122.39, 37.75$ |
| Straight Line Distance | 3705.441 | 3996.052 |
| Max distance b/w two locations | 15.09 | 13.34 |
| Max distance from home | 44.65 | 51.29 |
| Number of locations | 9358 | 9345 |
| Average Jump length | 0.370 | 0.399 |
| Radius of Gyration | 7.976 | 6.826 |
| Random Individual Entropy | 13.191 | 13.189 |
| Real Individual Entropy | 13.191 | 12.337 |

We use the SF taxi traces to show information that can be used to distinguish between two cabs associated with two unique cab identifiers. Tables III and IV outline some of the mobility metrics that we described earlier in the section. All the distance values are in kilometers. Home location here is calculated using the coordinates where a cab starts and ends its day in. Since we have pre-processed the dataset to only consider a single taxi-cab for our results, the random and real location entropy does not apply for our case. Information like mean square displacement, straight line distance, radius of gyration, average jump length and max distance from home provides insights into the vehicle's trajectory patterns. A higher number of locations when combined with GPS coordinates, can be used to identify the popularity of the cab service in a particular area as compared to an area with fewer locations visited.

TABLE IV
NUMBER OF VISITS MADE BY A SINGLE TAXI TO EACH OF THE FIVE LOCATIONS.

| Latitude | Longitude | N visits | Frequency Rank |
|---|---|---|---|
| -122.38602 | 37.61637 | 8 | 1 |
| -122.38616 | 37.61636 | 6 | 2 |
| -122.39472 | 37.78704 | 6 | 3 |
| -122.38613 | 37.61652 | 6 | 4 |
| -122.43461 | 37.78885 | 5 | 5 |

Some recent real life use cases of the SF taxi traces in the last couple of years are: traffic forecasting using a temporal directed graph convolution network by studying temporal tendencies and periodicities in movement characteristics of vehicles [189]; Urban hotspot area detection using spatial clustering, mined from the taxi trajectory data, which directly represents an user's travel characteristics and the operational status of urban traffic [190]; optimizations to taxi-sharing and ride-sharing mechanisms and algorithms [191]–[193]; Optimizing passenger finding recommendation algorithms for taxis that suffer from load balancing issue [194]; Extracting the social/ operational dynamics from taxi trips to study vehicle and passenger movement to and from its origin to improve road regulations and create new public transportation routes [195]; Optimizations to density blocking algorithms and trip demand merging strategies to propose an effective and scalable solution to the load-balancing problem [196]; studying historical trajectories to predict vehicle's next location [197]; Instantly discovering outlier trips from taxi trajectories [198].

TABLE V

COLLECTIVE MOBILITY METRICS FOR A SET OF LOCATIONS WITHIN THE BRIGHTKITE DATASET. RLE STANDS FOR RANDOM LOCATION ENTROPY, ULE STANDS FOR UNCORRELATED LOCATION ENTROPY AND N VISITS ARE NUMBER OF VISITS.

| Latitude | Longitude | RLE | ULE | N visits |
| --- | --- | --- | --- | --- |
| -122.419415 | 37.774929 | 3.000000 | 1.651086 | 682 |
| -122.346270 | 37.579938 | 2.807355 | 1.534141 | 387 |
| -122.389979 | 37.615223 | 2.584963 | 1.521993 | 375 |
| -122.343679 | 37.580304 | 2.584963 | 1.427061 | 303 |
| -122.346108 | 37.579361 | 2.321928 | 1.427061 | 250 |

We use the check-in information from the location based social networking application Brightkite, to showcase two different types of information user and location specific that can provide insights on travel patterns. Table V outlines some of the collective mobility metrics for a set of locations, and Tables VI and VII showcase individual mobility metrics for a set of 5 users, metrics as defined earlier in the section. We have processed several users for this dataset which provides insights that vary from our single user taxi trace analysis. The collective mobility metrics have similar location specific insights like the taxi traces, where they represent popularity of a location based on the number of visits made to it. The individual metrics can derive three types of information: individual user trajectory patterns, user travel choices and overlapping mobility patterns across users.

TABLE VI

INDIVIDUAL MOBILITY METRICS FOR A SET OF USERS WITHIN THE BRIGHTKITE DATASET. SLD STANDS FOR STRAIGHT LINE DISTANCE, MXDH STANDS FOR MAX DISTANCE FROM HOME AND MD, STANDS FOR MAXIMUM DISTANCE TRAVELLED.

| UID | SLD | Home Location | MXDH | MD |
| --- | --- | --- | --- | --- |
| 0 | 391897.399238 | $-105.06, 39.89$ | 12606.96 | 11517.62 |
| 1 | 773782.949699 | $-122.41, 37.63$ | 12800.55 | 12804.89 |
| 2 | 107469.649072 | $-104.98, 39.73$ | 12329.31 | 12332.04 |
| 3 | 476575.073371 | $-122.45, 37.74$ | 12799.75 | 12803.25 |
| 4 | 298975.023233 | $24.94, 60.18$ | 16035.99 | 16170.75 |

Some latest use cases for the Brightkite dataset include: Analyze role of LBSN check-ins using social community detection methods to extract city structured communities (SoLoMo cities) to eventually detect behavioral events changing city's communities [199]; Human mobility prediction approach using movement patterns with k-Latest Check-ins (kLC) [200]; Friend relationship judgment methods based on improved gravity models, using residence distance and spatial temporal co-occurrence zone as an influence on friendship judgment [201]; Graph models like reconstruction graph model with fusion feature, designed for mining potential social connections with the help of users' spatial information, that will ultimately reduce the negative effect caused by the sparsity of social connection graph [202]; Personalized recommendation tool solutions for suggesting interesting and new locations to users by bridging preference-aware and social-based recommendations [203]; Naıve Bayes Prediction Model derived using Bayesian Theory for point of interest recommendations [204]; Creating relationship-protection algorithms based on location-visiting characteristics [205]; Markov chain based position prediction model using multidimensional correction (MDC-MCM) [206]; Using advantages of regularity in human trajectories to model spatio-temporal information [207]; Identify social triad classes in a homophilic network to analyze the correlation between social triads and homophily [208].

TABLE VII

INDIVIDUAL MOBILITY METRICS FOR A SET OF USERS WITHIN THE BRIGHTKITE DATASET. UID STANDS FOR UNIQUE USER IDENTIFIER, N LOC STANDS FOR NUMBER OF LOCATIONS VISITED, N VISITS STANDS FOR NUMBER OF TOTAL VISITS MADE ACROSS ALL THE LOCATIONS, RG STANDS FOR RADIUS OF GYRATION, RIE STANDS FOR RANDOM INDIVIDUAL ENTROPY, REIE STANDS FOR REAL INDIVIDUAL ENTROPY.

| UID | N Loc | N visits | RG | RIE | REIE |
|---|---|---|---|---|---|
| 0 | 543 | 2100 | 2024.37 | 9.08 | 4.90 |
| 1 | 97 | 1210 | 2482.82 | 6.59 | 2.20 |
| 2 | 460 | 2100 | 2130.49 | 8.84 | 4.46 |
| 3 | 614 | 1807 | 1996.66 | 9.26 | 4.78 |
| 4 | 216 | 779 | 8207.31 | 7.75 | 3.58 |

### *E. Data privacy for open source datasets*

Open source mobility data at the point of collection almost always contains personal identifiable information (PII) like the user's name, contact information, trip history with exact location precision , payment information etc. Before sharing this information with government agencies or research groups, it is important for the mobility operators to mask or remove such information from the datasets. An example of a privacy specification developed by the Open Mobility Foundation [209] is the Mobility Data Specification (MDS) [210], [211], which makes available a set of APIs to make anonymized mobility data available as an open source resource. MDS is specifically used for location data derived from vehicles and provides information like trajectories, popular visit points, wait times etc. Since this data is collected by the vehicle and not the user device, there is no PII revealed. An important aspect of mobility data privacy is real-time data and its role in improving micro-mobility efficiency. Foundations that tackle mobility data privacy have collaborated with mobility operators to come up with standards and specifications to make such data available in anonymized or aggregated formats. Most of these standards come up with policies and non-infringement agreements that enforce good data-sharing practices without compromising on the privacy of PII.

Most of the open source datasets that we have accessed are from websites like CRAWDAD, which makes the users sign a nonexclusive, non transferable, data license agreement before getting access to any of the information, with the caveat that data is not redistributed [212]. The San Francisco taxi traces were part of CRAWDAD and so fall under their license. The Google Mobility Trend Report [145] follows the company's stringent privacy policy and provides information about the percentage rise and fall of Map requests to a given location, at no point making any PII available like an individual's location, contacts or movement. This dataset is derived from aggregated/anonymized sets of data from user's who have specifically turned on their Location History in the Google maps application. Google uses differential privacy to add noise to the datasets, which provides the same insights as real data without revealing any PII. The Stanford Brightkite dataset does not specifically talk about licenses, but the data is anonymized to the point where we can group a random set of user checkins based on similarity in check-in patterns, but we cannot identify who each user in the group is.

TABLE VIII
OPEN SOURCING AND PRIVACY POLICIES OF POPULAR UMA TRACES CLASSIFIED USING OUR TAXONOMY

| TRACES | SOURCE | LICENSE | ANONYMIZATION | STRUCTURE |
|---|---|---|---|---|
| T-Drive [47] | Kaggle | Attribution | Pseudo-anonymized | Non-aggregated |
| Crivello [35] | Unpublished | N/A | N/A | N/A |
| Apple Maps [48], [49] | data.world | Public Domain | Anonymized | Aggregated |
| Google Maps [49], [145] | Google | Public Domain | Anonymized | Aggregated |
| Descartes Lab [36] | GITHUB | Attribution | Anonymized | Aggregated |
| KCMD-DDH [50] | KCMD Data Portal | Public Domain | Anonymized | Aggregated |
| JRC(Europe) [53] | Urban Mobility Data Platform | Public Domain | Anonymized | Aggregated |
| GIM [146] | Flow Monitoring | Public Domain | Anonymized | Aggregated |
| Gowalla [55] | Snap Stanford | Attribution | Pseudo-anonymized | Non-aggregated |
| Brightkite [55] | Snap Stanford | Attribution | Pseudo-anonymized | Non-aggregated |
| Nsense [135] | Snap Stanford | Attribution | Pseudo-anonymized | Non-aggregated |
| Cabspotting [147] | CRAWDAD | Attribution | Pseudo-anonymized | Non-aggregated |
| Geolife [136] | Microsoft Research | Public Domain | Anonymized | Aggregated |
| NYC Mobility [148] | data.world | Attribution | Pseudo-anonymized | Non-aggregated |
| GRID Bikeshare [56] | data.world | Attribution | Pseudo-anonymized | Aggregated |
| Texas Mobility [149] | Austin Open Data portal | Public Domain | Anonymized | Aggregated |
| UILM [150] | data.world | Attribution | Pseudo-anonymized | Aggregated |
| GSMC [37] | CRAWDAD | Attribution | Pseudo-anonymized | Non-Aggregated |
| Flexran [38] | CRAWDAD | Attribution | Pseudo-anonymized | Non-Aggregated |
| KTH [39] | CRAWDAD | Attribution | Pseudo-anonymized | Non-Aggregated |
| BLEBeacon [40] | CRAWDAD | Attribution | Pseudo-anonymized | Non-Aggregated |
| HYCCUPS [41] | CRAWDAD | Attribution | Pseudo-anonymized | Non-Aggregated |
| Cambridge Haggle [42] | CRAWDAD | Attribution | Pseudo-anonymized | Non-Aggregated |
| Fire Dpt Asturius [43] | CRAWDAD | Attribution | Pseudo-anonymized | Aggregated |
| SocialBlueConn [44] | CRAWDAD | Attribution | Pseudo-anonymized | Non-Aggregated |
| Rome Taxis [51] | CRAWDAD | Attribution | Pseudo-anonymized | Non-Aggregated |
| SIGCOMM 2009 [45] | CRAWDAD | Attribution | Pseudo-anonymized | Aggregated |
| Commercial Seoul [46] | CRAWDAD | Attribution | Pseudo-anonymized | Aggregated |
| Chicago taxi [52] | Kaggle | Attribution | Pseudo-anonymized | Non-Aggregated |
| Pedestrian Louisville [151] | data.world | Public Domain | Anonymized | Aggregated |
| MHealthDroid [152] | UCI Machine Learning Repository | Attribution | Pseudo-anonymized | Non-Aggregated |
| NetAAL [153] | WNLAB | Attribution | Anonymized | Aggregated |

Table VIII discusses the different sources, licenses, anonymization and aggregation standards of the various uMA traces. The sources column outlines the websites where the datasets are hosted. Licenses address the type of permissions required for distribution and reuse of these datasets. The different types of data licenses available are [213]:

- Public Domain The dataset has been dedicated to the public by waiving all rights to the research data worldwide under copyright law, including all related and neighboring rights, to the extent allowed by law.
- Attribution Appropriate credit is given where necessary by providing a link to the license or citations, and indicating if changes were made.
- Share-alike Remixing, transforming, or building upon the material, must include distributing your contributions under the same license as the original.
- Non-commercial Cannot use the material for commercial purposes.
- Database Only License applies to the database only and not its contents or data.

• No Derivatives No Derivative Works. Cannot alter, transform, or build upon this work.

The anonymization column describes if the data is completely or partially anonymized. Complete anonymization is when the patterns in a dataset cannot be traced back to the original users under any circumstances, while pseudo anonymization means that even though the data is currently anonymized, they are ways to derive connections between the original user and the features, which can be a privacy concern in the case of datasets heavy on personal identifiable information (PII). The aggregation column suggests if a dataset is aggregated or not aggregated. An aggregated dataset like the Google covid dataset, will have derived collective information for a group of individual records, while a nonaggregated dataset like Cabspotting will have data points for each individual user.

## V. CONCLUSION

Motivated by a wide range of applications from urban planning, efficient communication, transit and transportation, public health and healthcare, commerce, critical network infrastructure planning and provisioning, to name a few, it has become increasingly important to better understand human mobility and activity. As a result, uMA characterization and modeling has been attracting considerable attention from researchers and practitioners. uMA records and traces have played a crucial role in enabling the exploration of how Humans move in a variety of environments.

In this paper, we survey the current state-of-the-art of publicly available uMA research. The main contributions of our work include: (1) Proposing a novel taxonomy that classifies these traces based on a number of factors including their mobility mode, data source, data collection technology, information type and their current and potential future applications; (2) Categorizing several well-known public uMA datasets using the proposed taxonomy, along with providing their published source and data sharing policies; and (3) Using three uMA traces, each uniquely categorized using our taxonomy, to show real application of our taxonomy. Our study also discusses significant challenges associated with the publication and availability of real uMA traces, which goes on to motivate our ongoing work on developing realistic uMA trace generators. As part of future work, we are developing a comprehensive framework to collect, store, generate and analyze uMA datasets. We are using the features identified in our taxonomy as knobs that can be selected to generate realistic uMA traces across all taxonomy categories with equally high fidelity.